\documentclass{article}

\usepackage{microtype}
\usepackage{graphicx}
\usepackage{subfigure}
\usepackage{booktabs} % for professional tables
\usepackage{array,booktabs,arydshln,xcolor}
\usepackage{graphicx,wrapfig,lipsum}
\usepackage{float}
\usepackage{natbib}

\usepackage{hyperref}
\usepackage{verbatim}
\usepackage{amssymb}
\usepackage{array}
\usepackage[ruled,linesnumbered,vlined]{algorithm2e}
\newcommand{\PreserveBackslash}[1]{\let\temp=\\#1\let\\=\temp}
\newcolumntype{C}[1]{>{\PreserveBackslash\centering}p{#1}}

\usepackage{tikz} 
\usetikzlibrary{automata, positioning, arrows}

\usepackage{amsmath}
\usepackage[noabbrev,capitalize]{cleveref}
\usepackage{bbm, dsfont}
\usepackage[accepted]{icml2020}
\allowdisplaybreaks

\usepackage{nccmath}
\usepackage{mathtools}
\usepackage{nicefrac} 
\usepackage{xfrac} 

\usepackage{lipsum}

\usepackage{multicol}
\usepackage{lipsum}
\usepackage{tikz}
\usetikzlibrary{positioning,angles,quotes,arrows,automata}

\usepackage{float}
\usepackage{bm}
\usepackage{amsfonts}
\usepackage{amsbsy}
\usepackage{amsthm}
\usepackage{color}
\usepackage{amsmath}
\usepackage{enumerate}% http://ctan.org/pkg/enumerate
\usepackage{amssymb}
\usepackage{enumitem}
\usepackage{array}
\usepackage{nicefrac}

\usepackage{tabularx}
\usepackage{booktabs}
\usepackage{multirow}
\newcolumntype{Y}{>{\centering\arraybackslash}X}

\usepackage{color, colortbl}
\definecolor{Gray}{gray}{0.98}
\definecolor{LightCyan}{rgb}{0.88,1,1}
\newcolumntype{g}{>{\columncolor{Gray}}c}

\DeclareMathOperator*{\argmax}{argmax}
\DeclarePairedDelimiter\abs{\lvert}{\rvert}%
\DeclarePairedDelimiter\norm{\lVert}{\rVert}%
\makeatletter
\let\oldabs\abs
\def\abs{\@ifstar{\oldabs}{\oldabs*}}
\let\oldnorm\norm
\def\norm{\@ifstar{\oldnorm}{\oldnorm*}}
\makeatother

\newcommand{\eat}[1]{}
\newcommand{\states}{\mathcal{S}}
\newcommand{\actions}{\mathcal{A}}

\renewcommand{\P}{\mathbb{P}}

\newcommand{\Real}{\mathbb{R}}

\newcommand{\real}{\mathbb{R}}

\newcommand{\simplexs}{\Delta ^{ \states }}

\newcommand{\opt}{^\star}

\usepackage{thmtools}
\usepackage{nameref}
\usepackage[noabbrev,capitalize]{cleveref}

\makeatletter
\newcommand{\myref}[1]{\cref{#1}\mynameref{#1}{\csname r@#1\endcsname}}
\newcommand{\Myref}[1]{\Cref{#1}\mynameref{#1}{\csname r@#1\endcsname}}

\def\mynameref#1#2{%
	\begingroup
	\edef\@mytxt{#2}%
	\edef\@mytst{\expandafter\@thirdoffive\@mytxt}%
	\ifx\@mytst\empty\else
	\space(\nameref{#1})\fi
	\endgroup
}
\makeatother

\DeclareMathOperator{\E}{\mathbb{E}}

\newcommand{\ret}[1]{ g^\theta{(#1)}}
\newcommand{\ro}[1]{ g{(#1)}}
\newcommand{\indicator}{\mathbbm{1}}

\title{Entropic Risk Constrained Soft-Robust Policy Optimization}
\author{
Reazul Hasan Russel \hspace{16pt} Bahram Behzadian \hspace{16pt} Marek Petrik\\ 
Department of Computer Science\\
University of New Hampshire\\
Durham, NH 03824 USA\\
{\tt rrussel}, {\tt bahram}, {\tt mpetrik} at {\tt cs.unh.edu}
}
\date{}

\begin{document}
\maketitle

\begin{abstract}
Having a perfect model to compute the optimal policy is often infeasible in reinforcement learning. It is important in high-stakes domains to quantify and manage risk induced by model uncertainties. \emph{Entropic risk measure} is an exponential utility-based convex risk measure that satisfies many reasonable properties. In this paper, we propose an entropic risk constrained policy gradient and actor-critic algorithms that are risk-averse to the model uncertainty. We demonstrate the usefulness of our algorithms on several problem domains.
\end{abstract}

\section{Introduction}
Reinforcement Learning~(RL) aims to learn how to map situations to actions in order to maximize the rewards accrued over the long run~\citep{sutton2018reinforcement, szepesvari2010algorithms}. Markov Decision Processes (MDPs) provide a functional framework to model RL problems~\citep{Bertsekas1996,Puterman2005}. In general, transition dynamics and rewards of MDPs are computed from limited and noisy samples. Which often makes it difficult to build a good model of the world. This results in policies that can fail catastrophically when deployed~\citep{Petrik2016a,Hanasusanto2013}. To mitigate the risk of failure in high-stakes domains, such as autonomous driving or robotic manipulation, it is important to account for the uncertainty about models. 

A common approach to computing policies that are reliable even with imprecise models is to use \emph{robust optimization}~\citep{Iyengar2005,Wiesemann2013}. This approach is simple and can be computationally effective~\citep{Ho2018}, but unfortunately too conservative~\citep{petrik2019beyond}. A class of methods that build on robust optimization but mitigate its conservativeness are described as \emph{epistemic risk aversion}~\citep{Eriksson2019} or \emph{soft robustness}~\citep{Ben-Tal2010,Derman2019}. These methods also estimate the range of possible models consistent with the observed data and then optimize a policy with respect to a risk metric across different models. In one early example of this approach, the \emph{percentile criterion} optimizes the value-at-risk (VaR) of the policy's performance with respect to uncertain model~\citep{Delage2009}.

%It is important for many problems to optimize some measure of risk along with the standard objective.
%Standard optimization in infinite horizon MDPs maximizes the expected sum of discounted rewards. 

It is important to distinguish between soft-robustness with respect to epistemic uncertainty, which we address in this work, and standard risk-averse MDPs. Risk-averse MDPs optimize a risk-sensitive objective that penalizes the variability in returns caused by stochastic transitions, also referred to as the \emph{aleatoric uncertainty}. Policy gradient and actor-critic algorithms to optimize risk-averse objective for MDPs have been developed  recently~\citep{Tamar2013,Chow2014} for several common risk-measures like Value-at-risk (VaR) and conditional value-at-risk (CVaR)~\citep{Rockafellar2000}. These methods do not consider model uncertainty and are very different from our work in several crucial aspects. The use of VaR and CVaR in sequential optimization is complicated because they are not dynamically-consistent and the optimal policy may need to be history-dependent~\cite{Ruszczynski2010}.

In this paper, we propose entropic risk constrained policy gradient and actor-critic algorithms under epistemic uncertainty within a Bayesian framework. Our choice of relying on the entropic risk measure is motivated by the fact that, this risk measure is convex and time-consistent. Our contributions in this paper are as follows: 1) We derive gradient update rule for the entropic risk constrained optimization with model uncertainty where the sampling based gradients are estimated from a Bayesian posterior. 2) We propose a trajectory-based policy gradient algorithm and actor-critic algorithm with function approximation.

The remainder of the paper is organized as follows: \cref{sec:framework} formally describes the MDP framework and entropic risk constrained objective. \cref{sec:entpo} derives the gradient update rules and presents the policy gradient and actor-critic algorithms. \cref{sec:experiment} presents empirical results on several problem domains. And we finally draw conclusions in \cref{sec:conclusion}.

\section{Framework} \label{sec:framework}
We consider an MDP model $\Upsilon$ with a finite number of states $\states = \{1, \ldots, S \}$ and finite number of actions $\actions = \{1, \ldots, A\}$. Every action $a \in \actions$ is available for the decision maker to take in every state $s \in \states$. After taking an action $a\in\actions$ in state $s\in\states$, the decision maker receives a reward $r_{s,a} \in \real$ and transitions to a next state $s'$ according to the \emph{true} and \emph{unknown} transition probability $p\opt_{s,a} \in \simplexs$. 
%We use $P\opt : \states\times\actions\to\simplexs$ to denote the transition kernel and $\*p_{s,a}$ to denote the vector of transition probabilities from state $s$ and action $a$. A stationary policy $\pi(\cdot|s)$ is a probability distribution over actions $a\in\actions$ given a state $s\in\states$. 
We parameterize a class of stationary randomized policies as $\pi(\cdot|s;\theta)$ where $s\in\states$ and $\theta\in\Theta\subseteq\Real^k$ is a $k$-dimensional parameter vector. We use $\pi$ and $\theta$ interchangeably for the rest of the paper. The return $\rho^\theta$ for a policy $\theta$ and a sampled trajectory $\xi$ is defined as: $g^\theta(\xi) =  \sum_{t=0}^{\infty} \gamma^t r_{s_t,\pi(s_t)}$~\citep{Puterman2005}, where $\xi = [s_0,a_0,\ldots]$. The expected values of the random variables $g^\theta(\xi)$ when $\xi$ starts from a specific state $s$ is defined as the value function of that state: $v^\theta(s) = \E\big[ g^\theta(\xi) \big]$. We can estimate the gradients of the return $\rho^\theta$ w.r.t the parameters $\theta$ from sampled trajectories $\xi$. The objective is then to maximize the infinite horizon $\gamma$-discounted return $\rho^\theta$ by adjusting the parameters $\theta$ in the direction of the gradients~\citep{sutton2018reinforcement}. Ideally, the optimal policy $\pi\opt \in \argmax_{\pi\in\Pi} \rho^\pi(P\opt)$ could be computed with a known $P\opt$, where $\Pi$ is the set of all stationary deterministic policies. This is impossible when the true transition probabilities $P\opt$ are unknown and only estimated from samples.

\paragraph{Entropic Risk Measure} Risk-averse methods address the challenge of computing a policy that is not too conservative in the worst-case scenario when $P\opt$ is unknown. The idea is to compute a policy that maximizes the expected return and satisfies a constraint that the worst-case return is above some preset threshold. Entropic risk measure $\rho : \mathbb{X} \rightarrow \Real$ is a popular risk measure based on exponential utility function and for a risk-aversion parameter $\alpha > 0$, it takes the form:

\begin{equation} \label{eq:defn_ent_risk}
    \rho_\alpha(X) = -\frac{1}{\alpha} \log \big( \E[e^{-\alpha X}] \big)
\end{equation}

The entropic risk measure defined in \cref{eq:defn_ent_risk} satisfies the properties of monotonicity, translation invariance and convexity, but does not satisfy the positive homogeneity property~\citep{FOllmer2011}. Similarly, we define the exponential utility based entropic Bellman operator as:

\begin{equation} \label{eq:bellman}
    T[v](s) = \max_{a\in \mathcal{A}} \Big[ -e^{-R(s,a)} + \gamma \sum_{s'\in S}P(s'|s,a)v(s') \Big]
\end{equation}

This entropic Bellman operator of (\ref{eq:bellman}) is a contraction and satisfies other standard properties. 
%see \cref{ax:ent_bellman} for details.

%\section{Entropic Risk Constrained Policy Gradient} \label{sec:ent_mdp_pg}
 
\paragraph{Soft-Robust Objective} We define the entropic-risk constrained soft-robust objective with the below optimization problem:
\vspace{-0.3in}
\begin{center}
\begin{equation} 
\begin{aligned} \label{eq:ent_constrained_obj}
&\max_\theta \E_\Upsilon \Big[ \E_\xi \big[ \ret{\xi} \big] \Big] \\
& \text{s.t. }
-\frac{1}{\alpha} \log \Big(\E_\Upsilon \big[ e^{-\alpha \E_\xi [\ret{\xi}] } \big]\Big) \ge \beta \end{aligned}    
\end{equation}
\end{center}

where $\beta \in \Real$ is the cost tolerance and $\E_\Upsilon$ represents the expectation with respect to different models. We assume that there exists a policy $\theta$ such that the optimization problem in (\ref{eq:ent_constrained_obj}) is feasible. The policy $\theta$ computed for this entropic-risk constrained MDP is history independent, thanks to the time-consistency property of entropic risk measure. 

We solve \cref{eq:ent_constrained_obj} by applying Lagrange relaxation procedure (see e.g. Chapter 3 of \cite{Bertsekas2003}), which turns it into an unconstrained optimization problem:

\begin{equation} \label{eq:ent_lagrange}
    \begin{aligned}
    \min_{\lambda \ge 0} &\max_\theta \Bigg( L(\theta,\lambda) = \sum_m P(m) \sum_\xi P_{\theta,m}(\xi) \ro{\xi}\\
    &+ \lambda \bigg( \sum_m P(m) e^{-\alpha \sum_\xi P_{\theta,m}(\xi) \ro{\xi}} - e^{-\alpha\beta} \bigg) \Bigg)
    \end{aligned}
\end{equation}

where $\lambda$ is the Lagrange multiplier. The goal here is to find a saddle point $(\theta^*,\lambda^*)$ that satisfies $L(\theta,\lambda^*) \ge L(\theta^*,\lambda^*) \ge L(\theta^*,\lambda)$, $\forall \theta$ and $\forall \lambda\ge0$. This is achieved by descending in $\theta$ and ascending in $\lambda$ using the gradients.

\section{Entropic Risk Constrained Policy Optimization} \label{sec:entpo}
We compute the gradient estimates of (\ref{eq:ent_lagrange}) with respect to $\theta$ and $\lambda$ to optimize the objective.

\begin{equation} \label{eq:grad_theta}
\begin{aligned}
    \nabla_\theta & L(\theta,\lambda) = \sum_m P(m) \sum_{\xi:P_{\theta,m}(\xi) \neq 0} \ro{\xi} P_{\theta,m}(\xi) \Big(1 - \\
    &\alpha\lambda e^{-\alpha \sum_{\xi:P_{\theta,m}(\xi) \neq 0} P_{\theta,m}(\xi) \ro{\xi}}\Big)
    \sum_{k=0}^{T-1} \frac{\nabla_\theta \pi_\theta(a_k | s_k)}{\pi_\theta(a_k | s_k)}
\end{aligned}
\end{equation}

\begin{equation} \label{eq:grad_lambda}
\nabla_\lambda L(\theta,\lambda) = \sum_m P(m) e^{-\alpha \sum_{\xi:P_\theta(\xi) \neq 0} P_{\theta,m}(\xi) \ro{\xi}} - e^{-\alpha\beta}
\end{equation}

See \cref{ax:gradient_ent} for the detailed derivation of the gradients. We use this gradient update rule to develop policy-gradient (PG) and actor-critic (AC) algorithms.

\begin{algorithm*} [!h]
	\KwIn{A differentiable policy parameterization $\pi(.|.,\theta)$, a differentiable state-value function parameterization $\hat{v}(s,w)$, confidence level $\alpha$, budget constraint $\beta$, model posterior $\mathcal{M}$ and initial state distribution $p_0$, step size schedules $\zeta_3$, $\zeta_2$ and $\zeta_1$.}
	\KwOut{Policy parameters $\theta$}
	Initialize actor parameters $\theta\gets\theta_0$, $\lambda\gets\lambda_0$ and critic parameter $w\gets w_0$;
	
	\For{$k\gets0,1,2,\ldots$}{
	    $\bar{\theta} \gets 0$, $\bar{\lambda} \gets 0$, $\bar{w} \gets 0$;
	    
	    \tcc{Computed expected gradient from M sampled MDPs}
    	\For{$m\gets0,1,2,\ldots,M$}{
	    Sample model from posterior: $\tilde{\Upsilon} \sim \mathcal{M}$;
	    
	    Sample initial state: $s_0 \sim p_0$;
	    
	    $\hat{\theta} \gets 0$, $\hat{\lambda}\gets 0$, $\hat{w}\gets 0$;
	    
	    \tcc{Simulate trajectories with current policy $\theta$ from sampled MDP}
		\For{$t\gets0,1,2,\ldots,T$}{
        Sample action: $a_t\sim \pi(\cdot|s_t,\theta)$;
        
        Observe reward $R(s_t,a_t)$ and next state $s_{t+1}\sim\tilde{\Upsilon}(\cdot|s_t,a_t)$;
        
        $\delta = \rho_\alpha\big( R(s_t,a_t) \big) + \hat{v}(s_t,w) - \hat{v}(s_{t+1},w)$ \tcp*{Compute TD error}
        
        $\hat{\theta} \gets \hat{\theta} + \delta \Big(1 - \alpha\lambda e^{-\alpha \delta}\Big) \frac{\nabla_\theta \pi_\theta(a_t | s_t)}{\pi_\theta(a_t | s_t)}$;
        
        $\hat{\lambda} \gets \hat{\lambda} + e^{-\alpha \delta} - e^{-\alpha\beta}$\;
        $\hat{w} \gets \hat{w} +  \delta \nabla_w \hat{v}(s_t,w)$;
        }
        $\bar{\theta} \gets \bar{\theta} + \hat{\theta}/T$, \hspace{0.1in}
        $\bar{\lambda} \gets \bar{\lambda} + \hat{\lambda}/T$, \hspace{0.1in} 
        $\bar{w} \gets \bar{w} + \hat{w}/T$;
        }
        \textbf{$\lambda$ update:} $\lambda \gets \lambda - \zeta_1(k)\bar{\lambda}/M$ \tcp*{Actor update}
        \textbf{$\theta$ update:} $\theta \gets \theta + \zeta_2(k) \bar{\theta}/M$ \tcp*{Actor update}
        \textbf{$w$ update:} $w\gets w+ \zeta_3(k) \bar{w}/M $ \tcp*{Critic update}
	}
	\Return $\theta$;
	\caption{Entropic Risk Constrained Soft-Robust Actor-Critic Algorithm}    \label{alg:actor_critic}
\end{algorithm*}

\paragraph{Policy gradient algorithm} At each episode, several MDPs are sampled from the posterior distribution. The PG method then updates its parameters $\theta$ and $\lambda$ based on the expected gradients estimated from several trajectories drawn from the sampled MDPs. \cref{alg:policy_gradient} in the \cref{ax:pg_algo} presents the pseudo-code of the policy gradient algorithm.

\vspace{-0.1in}
\paragraph{Actor-Critic algorithm}
The policy gradient algorithm proposed in \cref{alg:policy_gradient} has a very high variance. We address this issue by using bootstrapped function approximation. We propose a risk-averse incremental actor-critic algorithm that converges to a (\emph{local}) saddle point of the entropic risk constrained objective function $L(\theta,\lambda)$ defined in \cref{eq:ent_lagrange}. The gradient update rule for the actor-critic algorithm follows directly from \cref{eq:grad_theta} and \cref{eq:grad_lambda}. We use value function approximation to estimate the critic and update the parameters incrementally with expected gradient estimates. \cref{alg:actor_critic} presents the pseudo-code of the actor-critic algorithm.

\section{Empirical Evaluation} \label{sec:experiment}
In this section, we empirically evaluate policy gradient and actor critic methods on different problem domains. All the experiments are run with risk parameter $\alpha=0.9$ and budget constraint $\beta=1$ unless otherwise specified. We start with some samples $\mathcal{D}$ drawn by arbitrary baseline policies $\pi_b$ from the underlying true distribution $P^\star$. We then compute the Bayesian posterior from the prior $p$ and data $\mathcal{D}$. We use linear combination of features in all the experiments to approximate the value functions for critic.

\begin{figure} [h]
    \centering
    \begin{minipage}{0.45\textwidth}
        \includegraphics[width=0.99\linewidth]{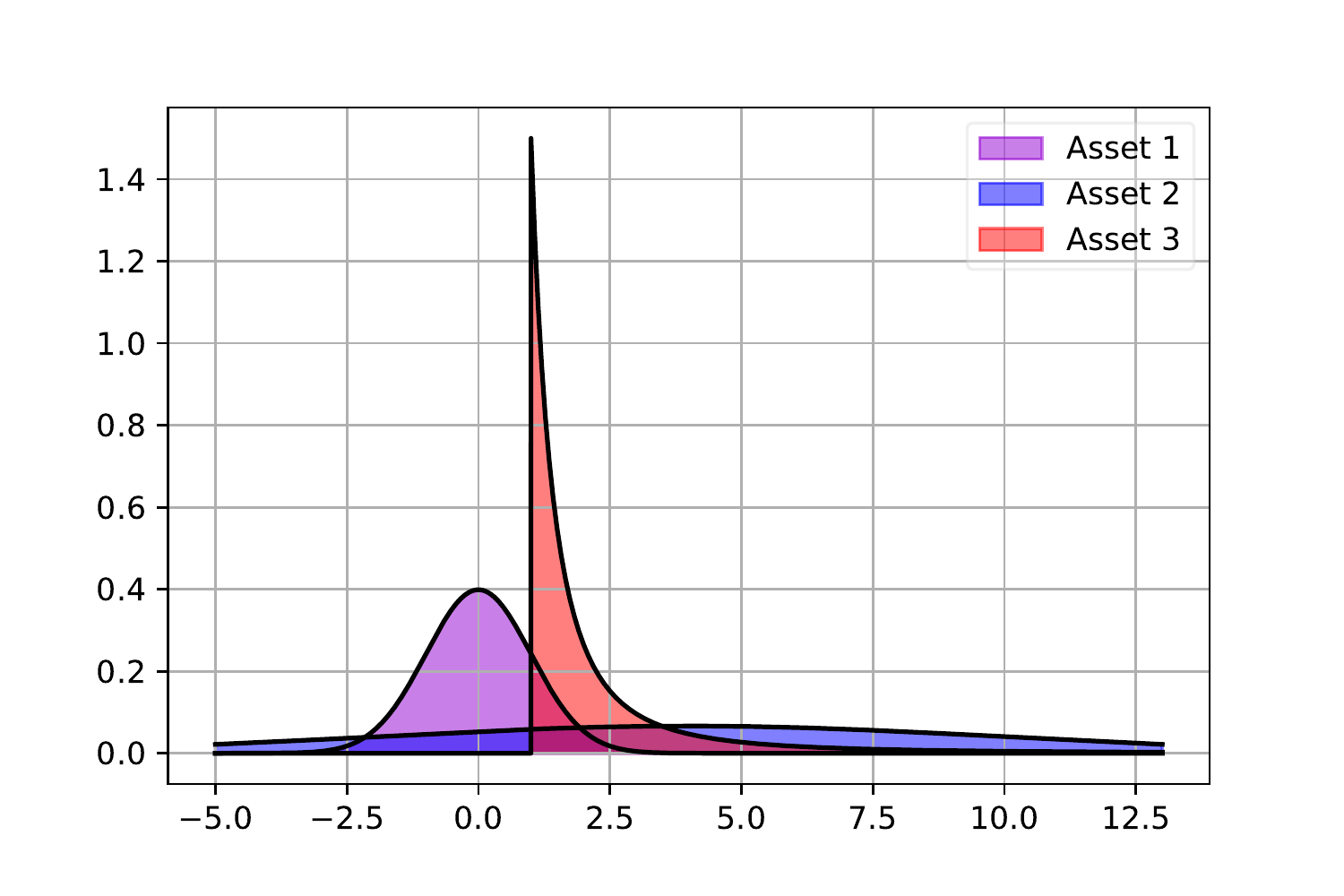}
    \end{minipage}
    \caption{Distribution of return for different assets.}
    \label{fig:ret_dist}
\end{figure}

\begin{figure*} [t]
    \centering
    \begin{minipage}{.45\textwidth}
        \includegraphics[width=0.99\linewidth]{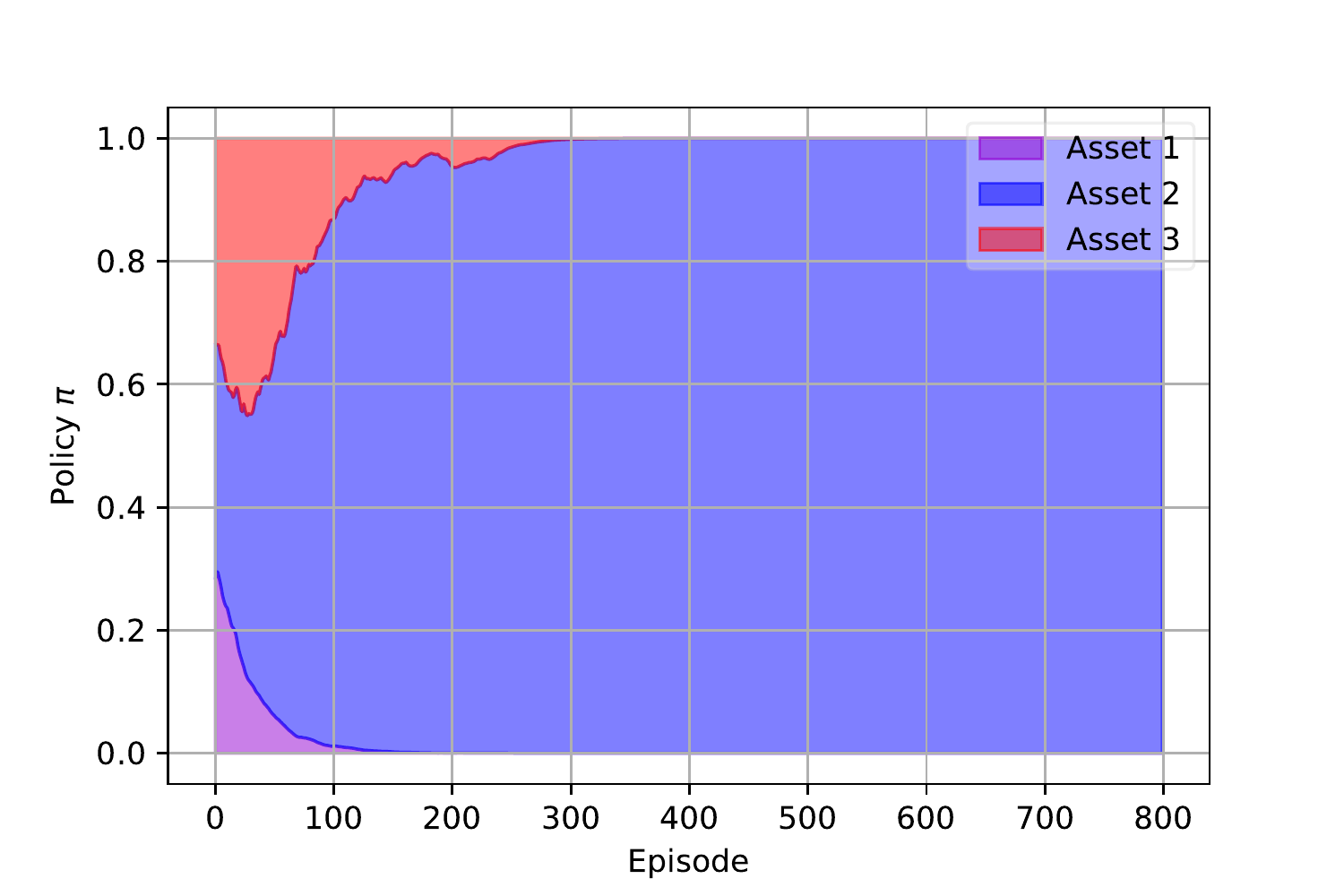}
    \end{minipage}%
    \begin{minipage}{0.45\textwidth}
        \includegraphics[width=0.99\linewidth]{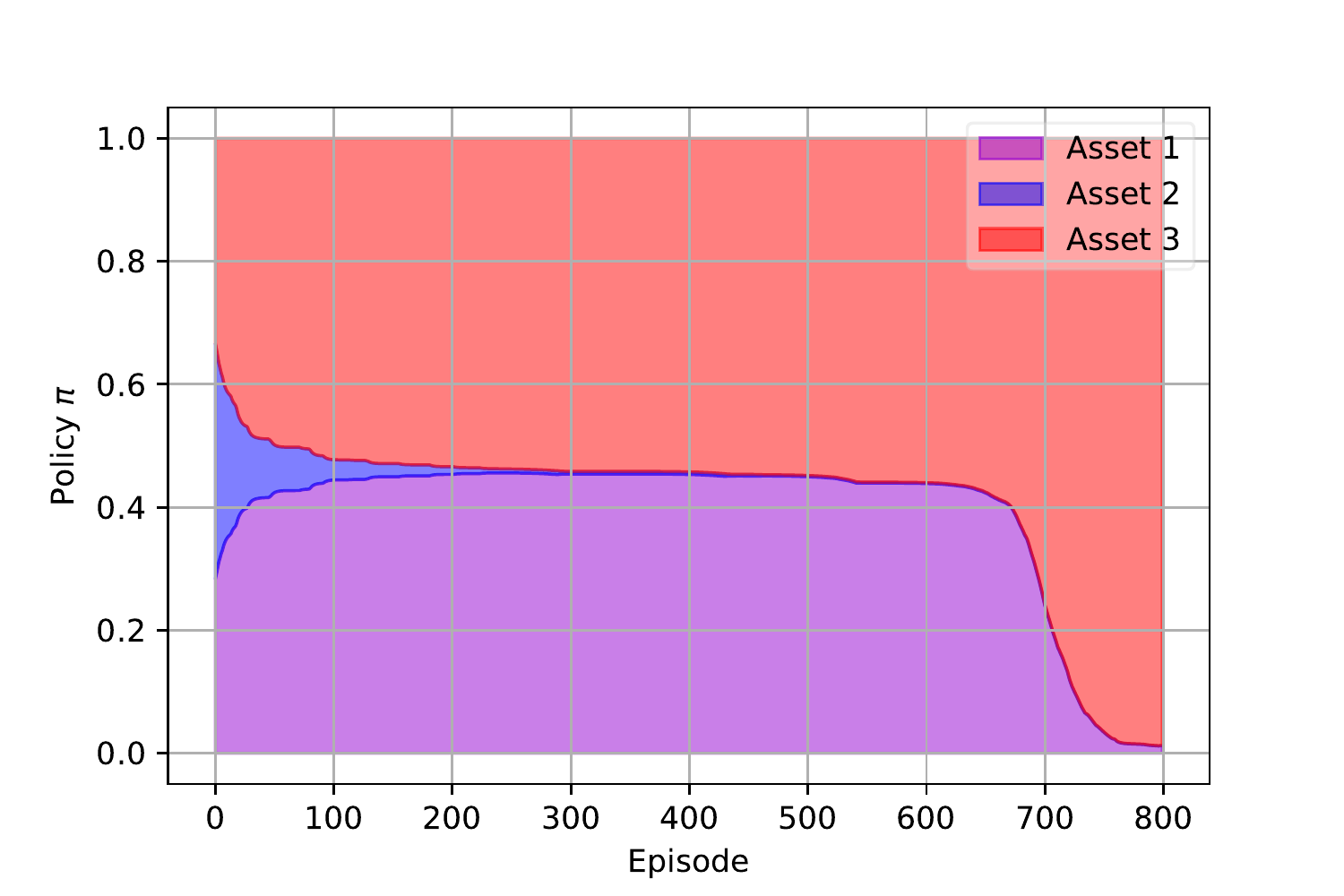}
    \end{minipage}
    \caption{Policy computed by policy gradient method for asset management problem, \textbf{left)} risk-neutral, \textbf{right)} soft-robust.}\label{fig:policy_asset}
\end{figure*}

\begin{figure*} [t]
\vskip -0.1in
    \centering
    \begin{minipage}{.45\textwidth}
        \includegraphics[width=0.99\linewidth]{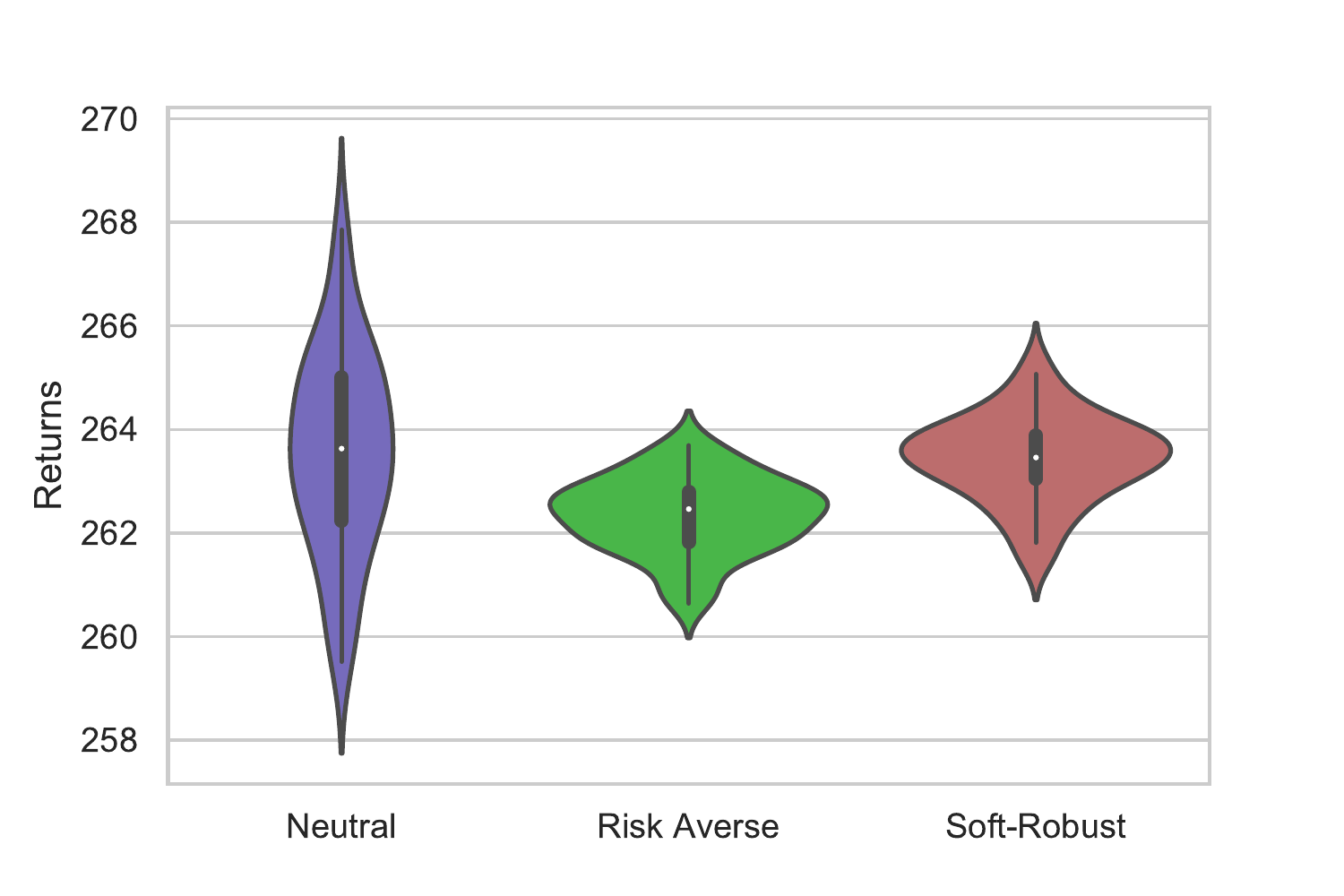}
    \end{minipage}%
    \begin{minipage}{0.45\textwidth}
        \includegraphics[width=0.99\linewidth]{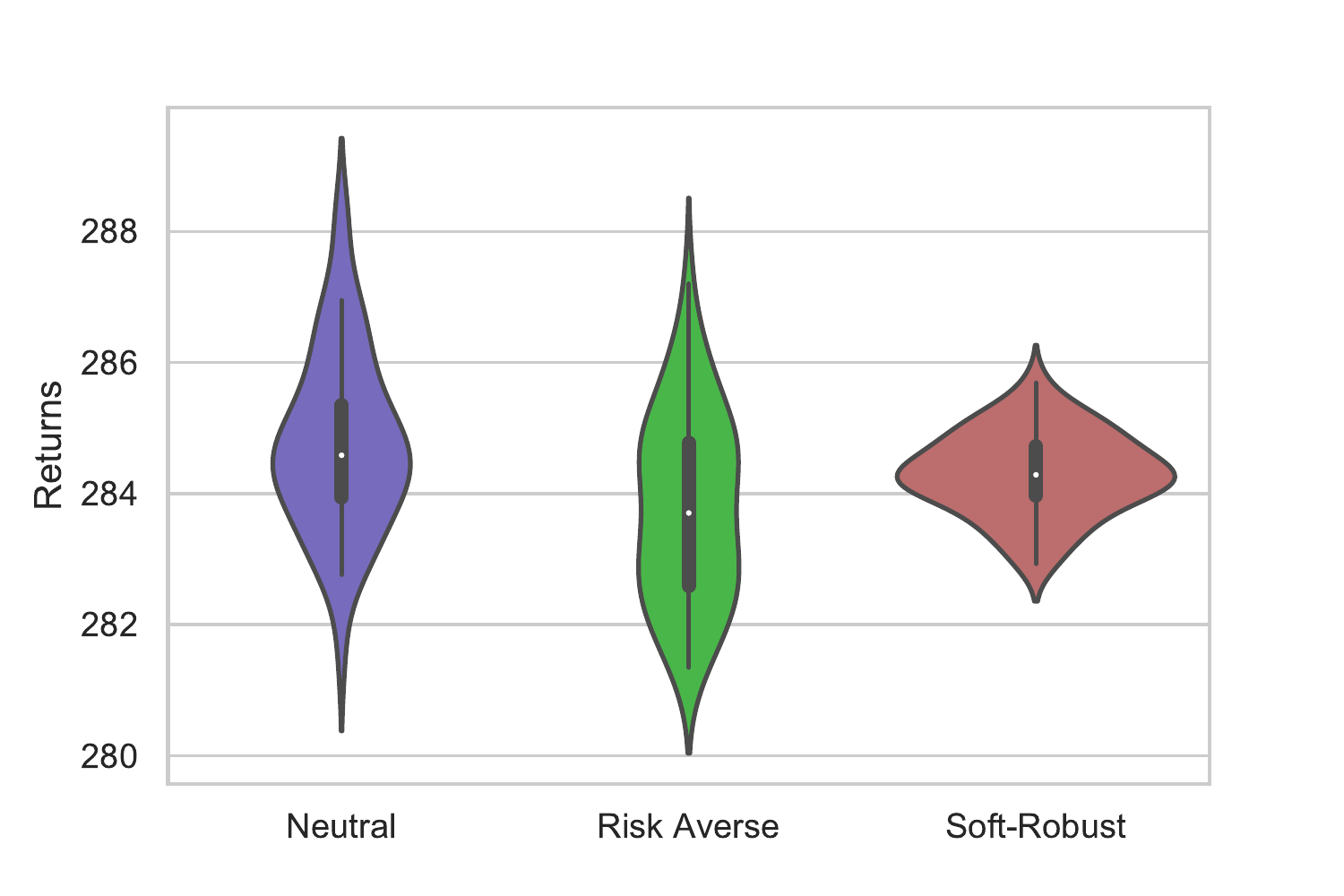}
    \end{minipage}
    \caption{Return distributions computed by actor-critic algorithms. \textbf{left)} inventory management problem, \textbf{right)} cart-pole problem.} \label{fig:ac_results}
\vspace{-0.1in}
\end{figure*}

\paragraph{Asset Management Problem} We first evaluate the policy gradient methods on a simple asset management problem~\citep{Tamar2015} with 3 assets, where the distribution of return for the first asset is standard normal. Asset 2 has a normal distribution with mean $\mu=4$ and standard deviation $\sigma=6$. Asset 3 has a pareto distribution with shape parameter $a=1.5$, scale parameter $m=1$ and pdf $p(x) = \frac{a m^a}{x^{a+1}}$. The pdf of the return distributions are shown in \cref{fig:ret_dist}. The outcome of an action is uncertain and that contributes to the uncertainty about model.

\cref{fig:policy_asset} shows the probability of picking each asset as the algorithm progresses. The risk neutral method on the left prioritizes asset 2 with higher mean return, avoiding the fact that it has high variance and the worst case return can be very bad. On the other hand, the soft-robust method on the right first avoids the most risky asset by allocating probabilities to less risky assets 1 and 3. It then realizes that asset 1 is riskier compared to asset 3 and thus allocates all the probabilities to asset 3.

\vspace{-0.15in}
\paragraph{Inventory Management} We now evaluate the actor-critic method on an instance of inventory management problem~\citep{Behzadian2019}, described in \cref{ax:invm}. The violin plot of \cref{fig:ac_results} (left) shows the return distributions computed by different actor-critic methods. The risk neutral method has very high variance and an arbitrarily bad worst-case return. As this domain involves both epistemic and aleatoric uncertainty, the performance of risk-averse and robust methods are competitive. The risk-averse method is able to reduce the variance due to the inherent stochasticity. Our soft-robust method has a slightly higher mean return and smaller variance compared to the risk-averse version.

\vspace{-0.1in}
\paragraph{Cart-Pole} Next, we evaluate our methods on Cart-Pole~\citep{openaigym}, a domain containing only epistemic uncertainty, details in \cref{ax:cartpole}. Return distributions computed by different actor-critic methods are shown in the violin plot of \cref{fig:ac_results} (right). The risk-averse actor-critic method performs very poorly having a variance as high as the risk-neutral case. This is because the environment dynamics of this domain are deterministic. But our soft-robust method obtains an expected return near to the risk-neutral method, and also reduces the variance by more than a factor of 2. The worst-case return estimate is also way higher compared to other methods.

\section{Summary and Conclusion} \label{sec:conclusion}
In this paper, we derived soft-robust gradient update rules for problems with model uncertainty. We proposed entropic risk constrained policy gradient and actor-critic algorithms with value function approximation. Our empirical results further establishes the usefulness of the proposed methods. Theoretical analysis of our algorithms remain to be done. Future work may also include the study of a novel class of algorithms that can be both risk-averse and soft-robust at the same time.

%We presented theoretical analysis for finite sample convergence guarantee of entropic risk measure. We also provided asymptotic convergence of our algorithms to local optima.
%Future work includes study of a novel class of algorithms that can be both robust (handling epsitemic uncertainty) and risk-averse (handling aleatoric uncertainty) at the same time and investigate their trade-offs.

\newpage

\bibliographystyle{icml2019}
\bibliography{bellman}

\begin{thebibliography}{23}
\providecommand{\natexlab}[1]{#1}
\providecommand{\url}[1]{\texttt{#1}}
\expandafter\ifx\csname urlstyle\endcsname\relax
  \providecommand{\doi}[1]{doi: #1}\else
  \providecommand{\doi}{doi: \begingroup \urlstyle{rm}\Url}\fi

\bibitem[Behzadian et~al.(2019)Behzadian, Russel, and Petrik]{Behzadian2019}
Behzadian, B., Russel, R.~H., and Petrik, M.
\newblock {High-Confidence Policy Optimization: Reshaping Ambiguity Sets in
  Robust MDPs}.
\newblock 2019.

\bibitem[Ben-Tal et~al.(2010)Ben-Tal, Bertsimas, and Brown]{Ben-Tal2010}
Ben-Tal, A., Bertsimas, D., and Brown, D.~B.
\newblock {A Soft Robust Model for Optimization Under Ambiguity}.
\newblock \emph{Operations Research}, 2010.

\bibitem[Bertsekas(2003)]{Bertsekas2003}
Bertsekas, D.~P.
\newblock \emph{{Nonlinear programming}}.
\newblock Athena Scientific, 2003.

\bibitem[Bertsekas \& Tsitsiklis(1996)Bertsekas and Tsitsiklis]{Bertsekas1996}
Bertsekas, D.~P. and Tsitsiklis, J.~N.
\newblock \emph{{Neuro-dynamic programming}}.
\newblock 1996.

\bibitem[Brockman et~al.(2016)Brockman, Cheung, Pettersson, Schneider,
  Schulman, Tang, and Zaremba]{openaigym}
Brockman, G., Cheung, V., Pettersson, L., Schneider, J., Schulman, J., Tang,
  J., and Zaremba, W.
\newblock Openai gym, 2016.

\bibitem[Chow \& Ghavamzadeh(2014)Chow and Ghavamzadeh]{Chow2014}
Chow, Y. and Ghavamzadeh, M.
\newblock {Algorithms for CVaR optimization in MDPs}.
\newblock \emph{Advances in Neural Information Processing Systems}, 2014.

\bibitem[Delage \& Mannor(2010)Delage and Mannor]{Delage2009}
Delage, E. and Mannor, S.
\newblock {Percentile Optimization for {Markov} Decision Processes with
  Parameter Uncertainty}.
\newblock \emph{Operations Research}, 2010.

\bibitem[Derman et~al.(2019)Derman, Mankowitz, Mann, and Mannor]{Derman2019}
Derman, E., Mankowitz, D., Mann, T., and Mannor, S.
\newblock {A {Bayesian} Approach to Robust Reinforcement Learning}.
\newblock \emph{Uncertainty in Artificial Intelligence (UAI)}, 2019.

\bibitem[Eriksson \& Christos(2019)Eriksson and Christos]{Eriksson2019}
Eriksson, H. and Christos, D.
\newblock {Epistemic Risk-Sensitive Reinforcement Learning}.
\newblock 2019.

\bibitem[F{\"{o}}llmer \& Knispel(2011)F{\"{o}}llmer and Knispel]{FOllmer2011}
F{\"{o}}llmer, H. and Knispel, T.
\newblock {Entropic Risk Measures: Coherence Vs. Convexity, Model Ambiguity and
  Robust Large Deviations}.
\newblock \emph{Stochastics and Dynamics}, 2011.

\bibitem[Hanasusanto \& Kuhn(2013)Hanasusanto and Kuhn]{Hanasusanto2013}
Hanasusanto, G. and Kuhn, D.
\newblock {Robust Data-Driven Dynamic Programming}.
\newblock In \emph{Advances in Neural Information Processing Systems (NIPS)},
  2013.

\bibitem[Ho et~al.(2018)Ho, Petrik, and Wiesemann]{Ho2018}
Ho, C.~P., Petrik, M., and Wiesemann, W.
\newblock {Fast Bellman Updates for Robust {MDP}s}.
\newblock \emph{Proceedings of Machine Learning Research (PMLR)}, 2018.

\bibitem[Iyengar(2005)]{Iyengar2005}
Iyengar, G.~N.
\newblock {Robust dynamic programming}.
\newblock \emph{Mathematics of Operations Research}, 2005.

\bibitem[Petrik et~al.(2016)Petrik, {Mohammad Ghavamzadeh}, and
  Chow]{Petrik2016a}
Petrik, M., {Mohammad Ghavamzadeh}, and Chow, Y.
\newblock {Safe Policy Improvement by Minimizing Robust Baseline Regret}.
\newblock \emph{Advances in Neural Information Processing Systems}, 2016.

\bibitem[Puterman(2005)]{Puterman2005}
Puterman, M.~L.
\newblock \emph{{{Markov} decision processes: Discrete stochastic dynamic
  programming}}.
\newblock John Wiley {\&} Sons, Inc., 2005.

\bibitem[Rockafellar \& Uryasev(2000)Rockafellar and Uryasev]{Rockafellar2000}
Rockafellar, R.~T. and Uryasev, S.
\newblock {Optimization of conditional value-at-risk}.
\newblock \emph{Journal of Risk}, 2000.

\bibitem[Russel \& Petrik(2019)Russel and Petrik]{petrik2019beyond}
Russel, R.~H. and Petrik, M.
\newblock Beyond confidence regions: Tight {Bayesian} ambiguity sets for robust
  {MDP}s.
\newblock \emph{Advances in Neural Information Processing Systems}, 2019.

\bibitem[Ruszczynski(2010)]{Ruszczynski2010}
Ruszczynski, A.
\newblock {Risk-averse dynamic programming for {Markov} decision processes}.
\newblock \emph{Mathematical Programming}, 2010.

\bibitem[Sutton \& Barto(2018)Sutton and Barto]{sutton2018reinforcement}
Sutton, R.~S. and Barto, A.~G.
\newblock \emph{Reinforcement learning: An introduction}.
\newblock MIT press, 2018.

\bibitem[Szepesv{\'a}ri(2010)]{szepesvari2010algorithms}
Szepesv{\'a}ri, C.
\newblock \emph{Algorithms for Reinforcement Learning}.
\newblock Morgan \& Claypool Publishers, 2010.

\bibitem[Tamar et~al.(2013)Tamar, Castro, and Mannor]{Tamar2013}
Tamar, A., Castro, D.~D., and Mannor, S.
\newblock {Temporal Difference Methods for the Variance of the Reward To Go}.
\newblock \emph{International Conference on Machine Learning}, 2013.

\bibitem[Tamar et~al.(2015)Tamar, Chow, Ghavamzadeh, and Mannor]{Tamar2015}
Tamar, A., Chow, Y., Ghavamzadeh, M., and Mannor, S.
\newblock {Policy Gradient for Coherent Risk Measures}.
\newblock \emph{Neural Information Processing Systems}, 2015.

\bibitem[Wiesemann et~al.(2013)Wiesemann, Kuhn, and Rustem]{Wiesemann2013}
Wiesemann, W., Kuhn, D., and Rustem, B.
\newblock {Robust {Markov} decision processes}.
\newblock \emph{Mathematics of Operations Research}, 2013.

\end{thebibliography}

\newpage
\appendix
\onecolumn

\section{Gradient Update Rule for Soft-Robust Objective} \label{ax:gradient_ent}

From \cref{eq:ent_lagrange}, we have:
\begin{equation*}
  L(\theta,\lambda) = \sum_m P(m) \sum_\xi P_{\theta,m}(\xi) \ro{\xi} + \lambda \bigg( \sum_m P(m) e^{-\alpha \sum_\xi P_{\theta,m}(\xi) \ro{\xi}} - e^{-\alpha\beta} \bigg)
\end{equation*}

First we compute the gradient of $L(\theta, \lambda)$ with respect to $\theta$.
\begin{center}
%\begin{equation*} \label{eq:entr_grad_theta}
\begingroup
\allowdisplaybreaks
\begin{align*}
\nabla_\theta L(\theta,\lambda) &= \sum_m P(m) \sum_{\xi:P_\theta(\xi) \neq 0} \nabla_\theta P_{\theta,m}(\xi) \ro{\xi} + \lambda \sum_m P(m) \nabla_\theta e^{-\alpha \sum_{\xi:\P_\theta(\xi) \neq 0} \P_{\theta,m}(\xi) \ro{\xi}} 
\\
&= \sum_m P(m) \Bigg( \sum_{\xi:P_\theta(\xi) \neq 0} \nabla_\theta P_{\theta,m}(\xi) \ro{\xi}\\ & \hspace{0.6in} + \lambda e^{-\alpha \sum_{\xi:P_\theta(\xi) \neq 0} P_{\theta,m}(\xi) \ro{\xi}} \nabla_\theta \Big( -\alpha \sum_{\xi:P_\theta(\xi) \neq 0} P_{\theta,m}(\xi) \ro{\xi} \Big) \Bigg)
\\
&= \sum_m P(m) \Bigg( \sum_{\xi:P_\theta(\xi) \neq 0} \nabla_\theta P_{\theta,m}(\xi) \ro{\xi} \\ &\hspace{0.9in} - \alpha\lambda e^{-\alpha \sum_{\xi:P_\theta(\xi) \neq 0} P_{\theta,m}(\xi) \ro{\xi}} \sum_{\xi:P_\theta(\xi) \neq 0} \nabla_\theta P_{\theta,m}(\xi) \ro{\xi} \Bigg)
\\
&= \sum_m P(m) \sum_{\xi:P_\theta(\xi) \neq 0} \nabla_\theta P_{\theta,m}(\xi) \ro{\xi} \Bigg( 1 - \alpha\lambda e^{-\alpha \sum_{\xi:P_\theta(\xi) \neq 0} P_{\theta,m}(\xi) \ro{\xi}} \Bigg)
\\
&= \sum_m P(m) \sum_{\xi:P_{\theta,m}(\xi) \neq 0} P_{\theta,m}(\xi) \nabla_\theta \log P_{\theta,m}(\xi) \ro{\xi}
\\ &\hspace{2.1in} \bigg( 1 - \alpha\lambda e^{-\alpha \sum_{\xi:P_{\theta,m}(\xi) \neq 0} P_{\theta,m}(\xi) \ro{\xi}} \bigg)
\\
&= \sum_m P(m) \sum_{\xi:P_{\theta,m}(\xi) \neq 0} \ro{\xi} P_{\theta,m}(\xi) \bigg( 1 - \alpha\lambda e^{-\alpha \sum_{\xi:P_{\theta,m}(\xi) \neq 0} P_{\theta,m}(\xi) \ro{\xi}} \bigg)
\\ & \hspace{1.2in} \nabla_\theta \log \Bigg( \prod_{k=0}^{T-1} P_m(s_{k+1 | s_k, a_k}) \pi_\theta(a_k | s_k) \indicator \{x_0 = x^0\} \Bigg)
\\
&= \sum_m P(m) \sum_{\xi:P_{\theta,m}(\xi) \neq 0} \ro{\xi} P_{\theta,m}(\xi) \bigg( 1 - \alpha\lambda e^{-\alpha \sum_{\xi:P_{\theta,m}(\xi) \neq 0} P_{\theta,m}(\xi) \ro{\xi}} \bigg) 
\\ & \hspace{0.5in} \nabla_\theta \Bigg( \sum_{k=0}^{T-1} \log P_m(s_{k+1 | s_k, a_k}) + \log \pi_\theta(a_k | s_k) + \log \indicator \{x_0 = x^0\} \Bigg)
\\
&= \sum_m P(m) \sum_{\xi:P_{\theta,m}(\xi) \neq 0} \ro{\xi} P_{\theta,m}(\xi) \bigg( 1 - \alpha\lambda e^{-\alpha \sum_{\xi:P_{\theta,m}(\xi) \neq 0} P_{\theta,m}(\xi) \ro{\xi}} \bigg)
\\ & \hspace{3.0in} \sum_{k=0}^{T-1} \nabla_\theta \log \pi_\theta(a_k | s_k)
\\
&= \sum_m P(m) \sum_{\xi:P_{\theta,m}(\xi) \neq 0} \ro{\xi} P_{\theta,m}(\xi) \bigg( 1 - \alpha\lambda e^{-\alpha \sum_{\xi:P_{\theta,m}(\xi) \neq 0} P_{\theta,m}(\xi) \ro{\xi}} \bigg)
\\ & \hspace{3.2in} \sum_{k=0}^{T-1} \frac{\nabla_\theta \pi_\theta(a_k | s_k)}{\pi_\theta(a_k | s_k)}
\end{align*}
\endgroup
%\end{equation*}
\end{center}

Next we compute the gradient of $L(\theta,\lambda)$ with respect to $\lambda$:
\begin{equation*} \label{eq:entr_grad_lambda}
\begin{aligned}
\nabla_\lambda L(\theta,\lambda) &= \sum_m P(m) e^{-\alpha \sum_{\xi:P_\theta(\xi) \neq 0} P_{\theta,m}(\xi) \ro{\xi}} - e^{-\alpha\beta}
\end{aligned}
\end{equation*}

\section{Policy Gradient Algorithm} \label{ax:pg_algo}
\begin{algorithm} [!h]
	\KwIn{A differentiable policy parameterization $\pi(.|.,\theta)$, confidence level $\alpha$, budget constraint $\beta$, model posterior $\mathcal{M}$ and initial state distribution $p_0$, step size schedules $\zeta_2$ and $\zeta_1$.}
	\KwOut{Policy parameters $\theta$}
	Initialize policy parameter $\theta\gets\theta_0$ and Lagrange parameter $\lambda\gets\lambda_0$;
	
	\For{$k\gets0,1,2,\ldots$}{%\While{TRUE}{
	$\hat{\theta}\gets 0$, $\hat{\lambda}\gets 0$;
	
	\tcc{Estimate gradients from M sampled MDPs}
    	\For{$m\gets0,1,2,\ldots,M$}{
    	    Sample model from posterior: $\tilde{\Upsilon} \sim \mathcal{M}$;
    	    
		    Sample initial state: $s_0 \sim p_0$;
		    
    		Generate trajectories for current policy $\theta$: $\Xi_\theta \sim \tilde{ \Upsilon}$;
            
            $\hat{\theta} \gets \hat{\theta} + \sum_{\xi\in\Xi_\theta} P_{\theta}(\xi) \ro{\xi} \Big(1 - \alpha\lambda e^{-\alpha \sum_{\xi\in\Xi_\theta} P_{\theta}(\xi) \ro{\xi}}\Big) \sum_{l=0}^{T-1} \frac{\nabla_\theta \pi_\theta(a_l | s_l)}{\pi_\theta(a_l | s_l)}$; 
            
            $\hat{\lambda} \gets \hat{\lambda} + \big( e^{-\alpha \sum_{\xi \in \Xi_\theta} P_{\theta}(\xi) \ro{\xi}} - e^{-\alpha\beta}$\big);
        }
        \tcc{Update parameters with expected gradient estimates}
        \textbf{$\theta$ update:} $\theta \gets \theta + \zeta_2(k) \hat{\theta}/M$;
        
        \textbf{$\lambda$ update:} $\lambda \gets \lambda - \zeta_1(k) \hat{\lambda}/M$;
	}
	\Return $\theta$;
	\caption{Entropic Risk Constrained Soft-Robust Policy Gradient Algorithm}    \label{alg:policy_gradient}
\end{algorithm}

\section{Experiment Details}

\subsection{Inventory Management} \label{ax:invm}
This is a full MDP setup with discrete state and action spaces. There is inherent stochasticity in transition dynamics between states and also the model parameters are not known precisely because of limited samples. So this domain involves both aleatoric and epsitemic uncertainty. It starts from an empty inventory level and the inventory evolves based on a normally distributed demand  $\sim\mathcal{N}(\mu=8,\sigma=3)$. The purchase cost and sale price are $2.49$ and $3.99$ respectively. Ordering products to restock the inventory helps to meet demands, but unsold products incur a holding cost of $0.03$. 

\subsection{Cart-pole} \label{ax:cartpole}
Cart-pole is a standard RL benchmark problem where the evolution of state space is deterministic (no aleatoric uncertainty). But the model parameters are not known precisely and the domain involves epistemic uncertainty. We build a linear model of transition dynamics with data-sets generated from the true distribution. We then generate synthetic samples from the fitted model and use K-nearest neighbor strategy to aggregate nearby states with a resolution of 200.

%\section{Algorithms} \label{alg:ax_algorithms}

\end{document}